\documentclass[11pt,a4paper]{article}
\usepackage[hyperref]{acl2017}
\usepackage{times}
\usepackage{latexsym}

\usepackage{url}

\usepackage{amsmath}
\usepackage{graphicx}

\aclfinalcopy 

\title{Relevance of Unsupervised Metrics in Task-Oriented Dialogue for Evaluating Natural Language Generation}

\author{Shikhar Sharma \and Layla El Asri \and Hannes Schulz \and Jeremie Zumer \\
        Microsoft Maluuba \\
        {\tt first.last@microsoft.com}}

\date{}

\begin{document}
\maketitle
\begin{abstract}
Automated metrics such as BLEU are widely used in the machine translation literature. They have also been used recently in the dialogue community for evaluating dialogue
response generation. However, previous work in dialogue response generation has shown that
these metrics do not correlate strongly with human judgment in the non task-oriented
dialogue setting. Task-oriented dialogue responses are expressed on narrower domains and exhibit lower diversity. It is thus reasonable to think that these automated metrics would correlate well with
human judgment in the task-oriented setting where the generation task consists of translating dialogue acts into a sentence. We conduct an empirical study to
confirm whether this is the case. Our findings indicate that these automated metrics
have stronger correlation with human judgments in the task-oriented setting compared
to what has been observed in the non task-oriented setting. We also observe that these metrics correlate even better for datasets which provide multiple ground truth reference sentences. In addition, we show that some of the currently available corpora for task-oriented language generation can be solved with simple models and advocate for more challenging datasets.
\end{abstract}

\section{Introduction}
Rule-based and template-based dialogue response generation systems have been around for
a long time~\cite{axelrod,rulebased1}. Even today, many
task-oriented dialogue systems deployed in production are
rule-based and template-based. These systems do not scale
with increasing domain complexity and maintaining the increasing
number of templates becomes cumbersome.
In the past, \newcite{stochasticnlgoh} proposed a corpus-based
approach for Natural Language Generation (NLG) for task-oriented
dialogue systems. Other statistical approaches were proposed using tree-based models and reinforcement learning
\cite{Walker:2007:IDA:1622637.1622648,rieser-lemon:2009:EACL}.
Recently, deep-learning based approaches
\cite{wensclstm15,sharma2017nlg,lowe2015incorporating,sordoni}
have shown promising results for dialogue response generation.

The automated evaluation of machine-generated language is challenging
and an important problem for the natural language processing community.
The most widely used automated metrics currently are word-overlap
based metrics such as BLEU~\cite{Papineni:02},
METEOR~\cite{Banerjee:05} which were proposed originally for machine
translation. While these metrics were shown to correlate well
with manual human evaluation in machine translation tasks,
previous studies showed that this is not the case in non-task oriented dialogue~\cite{LiuHowNotTo-D16-1230}. This is explained by the fact that for the same context (\textit{e.g.} a user utterance), responses in dialogue have more diversity.
Word-overlap metrics are unable to capture semantics and thus, can lead
to poor scores even for appropriate responses. Human evaluation in this case is
the most reliable metric. However, human judgments are expensive to
obtain and not readily available at all times.

Task-oriented dialogue systems are employed in
narrower domains (\textit{e.g.} booking a restaurant) and responses do not have as much diversity as
in the non-task oriented setting. Another important difference is that in the non-task oriented setting, response generation is often performed \textit{end-to-end}, which means that the model takes as input the last user utterance and potentially the dialogue history and it outputs the next system answer. In the task-oriented setting, on the other hand, the language generation task is often seen as a translation step from an abstract representation of a sentence to the sentence itself. As a consequence, automated metrics which compare a generated sentence to a reference sentence might be more appropriate and correlate with human judgments. In this paper, we:
\begin{itemize}
	\item study the correlation between human judgments and several unsupervised automated metrics on two popular task-oriented dialogue datasets,
	\item introduce variants of existing models and evaluate their performance on these metrics
\end{itemize}
We find that the automated metrics have stronger correlation with human judgments in the task-oriented setting than what has been observed in the non task-oriented setting. We also observe that these metrics correlate even more in the presence of multiple reference sentences.

\section{Related Work}
\newcite{LiuHowNotTo-D16-1230} did an empirical study to evaluate
the correlation between human scores and several automated
word-overlap metrics as well as embedding-based metrics for dialogue
response generation. They observed that these metrics, though widely used in the literature, had only weak correlation with human judgments in the non task-oriented dialogue NLG setting.

In terms of supervised NLG evaluation metrics, \newcite{adem} proposed the ADEM model which trains a hierarchical recurrent neural network in a supervised manner to predict human-like scores. This learned score was shown to correlate better with human judgments than any other automated metric. However, the drawback of this approach is the requirement for expensive human ratings.

\newcite{deeprldlg} proposed to use reinforcement learning to train an \textit{end-to-end} dialogue system. They simulate a dialogue between two agents and use a policy gradient algorithm with a reward function which evaluates specific properties of the responses generated by the dialogue system.

In the adversarial setting, \newcite{googleadversarial} train a recurrent neural network discriminator to differentiate human-generated responses from model-generated responses. However, an extensive analysis of the viability and the ease of standardization of this approach is yet to be conducted. \newcite{jiweiadversarial}, apart from adversarially training dialogue response models, propose an independent adversarial evaluation metric \textit{AdverSuc} and a measure of the model's reliability called \textit{evaluator reliability error}. Drawbacks of these approaches are that they are model-dependent.
Adversarial methods might be promising for task-oriented dialogue systems but more research needs to be conducted on their account. 

Most of the work described so far has been done in the non task-oriented dialogue setting as there has been prior work indicating that automated metrics do not correlate well with humans in that setting. There has not yet been any empirical validation that these conclusions also apply to the task oriented setting. Research in the task oriented setting has mostly made use of automated metrics such as BLEU and human evaluation \cite{wensclstm15,sharma2017nlg,ondrejdusek}.

\section{Metrics}
This section describes the set of automatic metrics whose correlation with human evaluation is studied. We consider first word-overlap metrics and then embedding-based metrics. In all that follows, when multiple references are provided, we compute the similarity between the prediction and all the references one-by-one, and then select the maximum value. We then average the scores across the entire corpus.
\subsection{Word-overlap based metrics}
\subsubsection{BLEU}
The BLEU metric \cite{Papineni:02} compares $n$-grams between the candidate utterance and the reference utterance. The BLEU score is computed at the corpus-level and relies on the following modified precision:
\begin{align}
p_n & = \nonumber \\
& \frac{\displaystyle\sum_{C \in \{Candidates\}} \displaystyle\sum_{n-gram \in C} Ct_{clip}(n-gram)}{\displaystyle\sum_{C' \in \{Candidates'\}} \displaystyle\sum_{n-gram' \in C'} Ct_{clip}(n-gram')} 
\end{align}
where \{Candidates\} are the candidate answers generated by the model and $Ct_{clip}$ is the clipped count for the $n$-gram which is the number of times the $n$-gram is common to the candidate answer and the reference answer clipped by the maximum number of occurrences of the $n$-gram in the reference answer.
The BLEU-N score is defined as:
\begin{align}
\text{BLEU-N} &= \text{BP}~\exp(\sum_n^N \omega_n \log(p_n))
\end{align}
where $N$ is the maximum length of the $n$-grams (in this paper, we compute BLEU-1 to BLEU-4), $\omega$ is a weighting that is often uniform and $\text{BP}$ is a brevity penalty. In this paper we report the BLEU score at the corpus level but we also compute this score at the sentence level to analyze its correlation with human evaluation.

\subsubsection{METEOR}
The METEOR metric \cite{Banerjee:05} was proposed as a metric which correlates better at the sentence level with human evaluation. To compute the METEOR score, first, an alignment between the candidate and the reference sentences is created by mapping each unigram in the candidate sentence to 0 or 1 unigram in the reference sentence. The alignment is not only based on exact matches but also stem, synonym, and paraphrase matches. Based on this alignment, unigram precision and recall are computed and the METEOR score is:
\begin{align}
\text{METEOR} &= F_{mean}(1 - p)
\end{align}
where $F_{mean}$ is the harmonic mean between precision and recall with the weight for recall 9 times a high as the weight for precision, and $p$ is a penalty.

\subsubsection{ROUGE}
ROUGE \cite{Lin:2004} is a set of metrics that was first introduced for summarization. We compute ROUGE-L which is an F-measure based on the Longest Common Subsequence (LCS) between the candidate and reference utterances.

\subsection{Embedding based metrics}
We consider another set of metrics which compute the cosine similarity between the embeddings of the predicted and the reference sentence instead of relying on word overlaps. 

\subsubsection{Skip-Thought}
The Skip-Thought model~\cite{skipthought} is trained in an unsupervised fashion and uses a recurrent network to encode a given sentence into an embedding and then decode it to predict the preceding and following sentences. The model was trained on the BookCorpus dataset~\cite{Zhu_2015_ICCV}. The embeddings produced by the encoder have a robust performance on semantic relatedness tasks. We use the pre-trained Skip-Thought encoder provided by the authors\footnote{https://github.com/ryankiros/skip-thoughts}.

We also compute other embedding-based methods which have been used as evaluation metrics for measuring human correlation in recent literature~\cite{LiuHowNotTo-D16-1230} for non task-oriented dialogue in Sections~\ref{subsubsec:embeddingavg}, \ref{subsubsec:embeddingextrema}, and \ref{subsubsec:embeddinggreedy}.

\subsubsection{Embedding average}
\label{subsubsec:embeddingavg}
This metric computes a sentence-level embedding by averaging the embeddings of the words composing this sentence:
\begin{align*}
\bar{e}_C &= \frac{\sum_{w \in C} e_w}{|\sum_{w' \in C} e_{w'}|}.
\end{align*}
In this equation, the vectors $e_w$ are embeddings for the words $w$ in the candidate sentence $C$.

\subsubsection{Vector extrema}
\label{subsubsec:embeddingextrema}
Vector extrema \cite{Forgues:14} computes a sentence-level embedding by taking the most extreme value of the embeddings of the words composing the sentence for each dimension of the embedding:
\begin{align*}
e_{rd} &= \begin{cases}
\max_{w \in C} e_{wd} & \text{if } e_{wd} > |\min_{w' \in C} e_{w'd}|\\
\min_{w \in C} e_{wd} & \text{otherwise}.
\end{cases}
\end{align*}
In this equation, $d$ is an index over the dimensions of the embedding and $C$ is the candidate sentence.

\subsubsection{Greedy matching}
\label{subsubsec:embeddinggreedy}
Greedy matching does not compute a sentence embedding but directly a similarity score between a candidate $C$ and a reference $r$ \cite{Rus:12}. This similarity score is computed as follows:
\begin{align*}
G(C, r) &= \frac{\sum{w \in C} \max_{\hat{w} \in r} cos\_sim(e_w, w_{\hat{w}})}{|C|}
\end{align*}
\begin{align}
GM(C, r) &= \frac{G(C, r) + G(r, C)}{2}.
\end{align}
In other words, each word in the candidate sentence is greedily matched to a word in the reference sentence based on the cosine similarity of their embeddings. The score is an average of these similarities over the number of words in the candidate sentence. The same score is computed by reversing the roles of the candidate and reference sentences and the average of the two scores gives the final similarity score.

\section{Response Generation Models}
\label{sec:models}
This section presents the different natural language generation models that we use in this study. All of these models take as input a set of dialogue acts \cite{Austin:62} potentially with slot types and slot values and translate this input into an utterance. An example input is \texttt{inform(food = Chinese)} and a corresponding output would be ``I am looking for a Chinese restaurant.''. In this example, the dialogue act is \texttt{inform}, the slot type is \textit{food}, and the slot value is \textit{Chinese}.
\begin{table*}
	\small
	\centering
	\begin{tabular}{|c|*{6}{c|}|*{6}{c|}}
		\hline
		& \multicolumn{5}{|c}{DSTC2} & \multicolumn{1}{|c||}{} & \multicolumn{6}{|c|}{Restaurants} \\ \cline{2-13}
		& B-1 & B-2 & B-3 & B-4 & M & R\_L & B-1 & B-2 & B-3 & B-4 & M & R\_L\\ \hline
		Gold & \bf 1.00 & \bf 1.00 & \bf 1.00 & \bf 1.00 & \bf 1.00 & \bf 1.00 & \bf 1.00 & \bf 1.00 & \bf 1.00 & \bf 1.00 & \bf 1.00 & \bf 1.00 \\ \hline \hline
		Random & 0.875 & 0.843 & 0.822 & 0.807 & 0.564 & 0.852 & 0.872 & 0.813 & 0.765 & 0.721 & 0.504 & 0.796 \\ \hline
		LSTM & 0.900 & 0.879 & 0.863 & 0.851 & 0.610 & 0.888 & 0.982 & 0.966 & 0.949 & 0.932 & 0.652 & 0.944 \\ \hline
		d-scLSTM & 0.880 & 0.850 & 0.828 & 0.812 & 0.578 & 0.874 & 0.980 & 0.964 & 0.948 & 0.931 & 0.654 & 0.945 \\ \hline
		hld-scLSTM & \bf 0.909 & \bf 0.890 & \bf 0.878 & \bf 0.870 & \bf 0.624 & \bf 0.899 & \bf 0.985 & \bf 0.978 & \bf 0.970 & \bf 0.962 & \bf 0.704 & \bf 0.965 \\ \hline
	\end{tabular}
	\caption{Performance comparison across models on word-overlap based automated metrics}
	\label{table:word-overlap}
\end{table*}
\begin{table*}
	\small
	\centering
	\begin{tabular}{|c|*{4}{c|}|*{4}{c|}}
		\hline
		& \multicolumn{3}{|c}{DSTC2} & \multicolumn{1}{|c||}{} & \multicolumn{4}{|c|}{Restaurants} \\ \cline{2-9}
		& Skip & Embedding & Vector & Greedy & Skip & Embedding & Vector & Greedy \\
		& Thought & Average & Extrema & Matching & Thought & Average & Extrema & Matching \\ \hline
		Gold & \bf 1.00 & \bf 1.00 & \bf 1.00 & \bf 1.00 & \bf 1.00 & \bf 1.00 & \bf 1.00 & \bf 1.00 \\ \hline \hline
		Random & 0.906 & 0.981 & 0.910 & 0.947 & 0.843 & 0.957 & 0.905 & 0.930 \\ \hline
		LSTM & \bf 0.946 & 0.985 & 0.935 & 0.962 & 0.945 & \bf 0.997 & 0.986 & 0.991 \\ \hline
		d-scLSTM & 0.925 & 0.984 & 0.926 & 0.957 & 0.948 & \bf 0.997 & 0.986 & 0.991 \\ \hline
		hld-scLSTM & 0.932 & \bf 0.987 & \bf 0.942 & \bf 0.964 & \bf 0.968 & \bf 0.997 & \bf 0.989 & \bf 0.993 \\ \hline
	\end{tabular}
	\caption{Performance comparison across models on sentence-embedding based automated metrics}
	\vspace*{-2mm}
	\label{table:embedding}
\end{table*}

\subsection{Random}
\label{subsec:random-model}
Given a dialogue act with one or more slot types, the random model finds all the examples in the training set with the same dialogue act
and slots (while ignoring slot values) and it randomly selects its output from this set of reference sentences. The datasets that we experiment on have some special slot
values such as ``yes'', ``no'', and ``don't care''. Since the model ignores all slot values, these special cases are not properly handled, which results in slightly lower performance than what we could get by spending more time hand-engineering the model's behavior for these values.

\subsection{LSTM}
\label{subsec:lstm-model}
This model consists of a recurrent LSTM~\cite{lstm} decoder. The dialogue acts and slot types are first encoded as a binary vector whose length is the number of possible combinations of dialogue acts and slot types in the dataset. We refer to this binary vector as the Dialogue Act ($DA$) vector. The $DA$
vector for a given set of dialogue acts is a binary vector over the fused dialogue act-slot types, e.g., \texttt{INFORM-FOOD}, \texttt{INFORM-COUNT}, etc.

This binary vector is given as input to the decoder at each time-step of the LSTM. The decoder then outputs a delexicalized sentence. A delexicalized sentence contains placeholders for the slot values. An example is ``I am looking for a FOOD restaurant.''. The values for the delexicalized slots (the type of food in this example) are then directly copied from the input. 

\subsection{delex-sc-LSTM}
\label{subsec:d-sclstm-model}
This model uses the same architecture as the LSTM model presented in the previous section except that it uses
sc-LSTM~\cite{wensclstm15} units in the decoder instead of LSTM units. We call this model the ``delex-sc-LSTM''\footnote{We will also refer to it as ``d-scLSTM''.}. As in the previous model, the input $DA$ vector only encodes acts and delexicalized slots. It does not contain any information about the slot value. 

By providing this model the same $DA$ vector input as the one given to the LSTM model, we can directly study if the additional
complexity of the sc-LSTM unit's reading gate provides significant improvement over the small-sized task-oriented dialogue datasets which are currently available.

\subsection{hierarchical-lex-delex-sc-LSTM}
\label{subsec:hld-sclstm-model}
This model is a variant of the ``ld-sc-LSTM'' model proposed by \newcite{sharma2017nlg} which is based on an encoder-decoder framework. We call our model ``hierarchical-lex-delex-sc-LSTM''\footnote{We will also refer to it as``hld-scLSTM''.}.
\begin{figure}[!h]
	\small
	\centering
	\includegraphics[width=\linewidth]{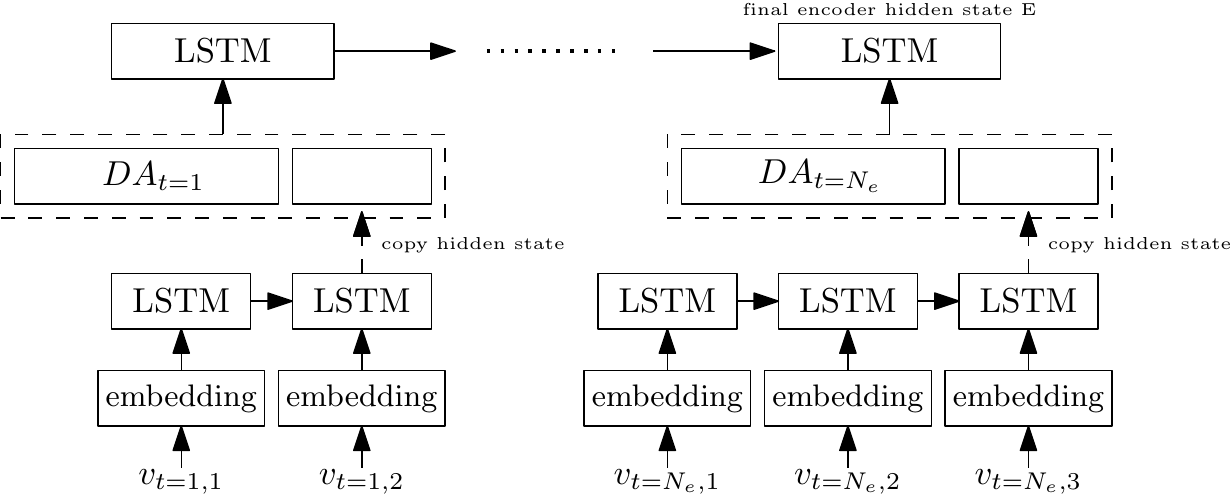}
	\caption{Encoder of the hld-scLSTM model}
	\label{fig:encoder}
\end{figure}
\begin{figure}[!h]
	\small
	\centering
	\includegraphics[width=\linewidth]{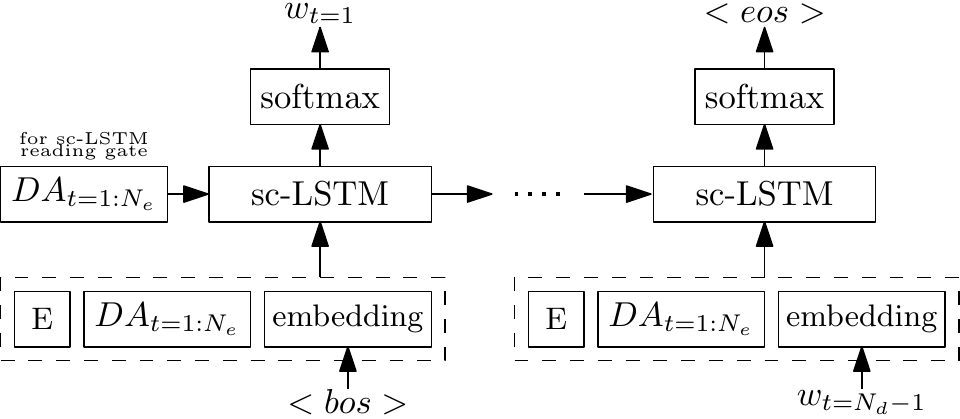}
	\caption{Decoder of the hld-scLSTM model}
	\label{fig:decoder}
\end{figure}

We present the encoder in Figure~\ref{fig:encoder}. The encoder consists of a hierarchical LSTM with $N_e$ time-steps, where $N_e$ is the
number of non-zero entries in the $DA$ vector. Each time-step of the encoder encodes
one dialogue act's delexicalized and lexicalized slot-value pair (e.g. (INFORM-FOOD,
`Chinese')). The delexicalized act-slot part is encoded as a one-hot vector which we refer to as $DA_t$. $DA_t$ is constructed by masking all except the
$t$\textsuperscript{th} dialogue act in the $DA$ vector\footnote{also referred to as
	$DA_{t=1:N_e}$}. The lexicalized value part is encoded by an LSTM encoder which
shares parameters across all time-steps and operates over the word-embeddings of the
lexicalized values $v_{t,i}$. Our model differs from the ``ld-sc-LSTM'' model in that we
use an LSTM encoder over the word-embeddings instead of computing the mean of the
word-embeddings. The final hidden state of this LSTM is concatenated with $DA_t$
and is given as input to the upper LSTM (see Figure~\ref{fig:encoder}). The final hidden state of the upper LSTM is then provided
to the decoder as input. This is another difference from the ``ld-sc-LSTM'' as that uses the mean of all the hidden states of the encoder instead,
which, in our experiments, did not perform as well as using just the final hidden state $E$. 

The decoder is described in
Figure~\ref{fig:decoder}. It is the same as in the ``ld-sc-LSTM'' model. At each
time-step, it takes as input the encoder output $E$, the $DA$ vector, and the
word-embedding of the word generated at the previous time-step. The $DA$ vector is
also additionally provided to the sc-LSTM cell in order for it to be regulated by its reading gate as described in \newcite{wensclstm15}.

\section{Experiments}
\begin{table*}
	\small
	\centering
	\begin{tabular}{|c|*{4}{c|}|*{4}{c|}}
		\hline
		& \multicolumn{3}{|c}{DSTC2} & \multicolumn{1}{|c||}{} & \multicolumn{4}{|c|}{Restaurants} \\ \cline{2-9}
		Metric & Spearman & p-value & Pearson & p-value & Spearman & p-value & Pearson & p-value \\ \hline
		Bleu 1             & -0.317 & \textless 0.005 & 0.583 & \textless 0.005
		& 0.069 & 0.494 & 0.277 & 0.005 \\
		Bleu 2             & -0.318 & \textless 0.005 & 0.526 & \textless 0.005
		& 0.091 & 0.366 & 0.166 & 0.099 \\
		Bleu 3             & -0.318 & \textless 0.005 & 0.500 & \textless 0.005
		& 0.109 & 0.280 & 0.223 & 0.026 \\
		Bleu 4             & -0.318 & \textless 0.005 & 0.461 & \textless 0.005
		& 0.105 & 0.296 & 0.255 & 0.010  \\
		METEOR             & 0.295 & \textless 0.005 & 0.582 & \textless 0.005
		& 0.353 & \textless 0.005 & 0.489 & \textless 0.005 \\
		ROUGE\_L           & 0.294 & \textless 0.005 & 0.448 & \textless 0.005
		& 0.346 & \textless 0.005 & 0.382 & \textless 0.005 \\
		Skip Thought       & 0.528 & \textless 0.005 & 0.086 & 0.397
		& 0.284 & \textless 0.005 & 0.364 & \textless 0.005 \\
		Embedding Average  & 0.295 & \textless 0.005 & 0.485 & \textless 0.005
		& 0.423 & \textless 0.005 & 0.260 & 0.009 \\
		Vector Extrema     & 0.299 & \textless 0.005 & 0.624 & \textless 0.005
		& 0.446 & \textless 0.005 & 0.287 & \textless 0.005 \\
		Greedy Matching    & 0.295 & \textless 0.005 & 0.572 & \textless 0.005
		& 0.446 & \textless 0.005 & 0.325 & \textless 0.005 \\ \hline \hline
		Human              & 0.810 & \textless 0.005 & 0.984 & \textless 0.005
		& 0.653 & \textless 0.005 & 0.857 & \textless 0.005 \\ \hline
	\end{tabular}
	\caption{Correlation of automated metrics with human evaluations scores}
	\label{table:correlation}
\end{table*}

\subsection{Decoding}
During training, at each time-step, we use the ground truth word
from the previous time-step. The model thus learns to generate the next word given the previous one. On the other hand, to generate sentences during test time, we use beam search. The first word input to the generator
is a special token $<bos>$ which indicates the beginning of the sequence. Decoding is stopped
if we reach a specified maximum number of time-steps or if the model outputs a
special token $<eos>$ which indicates the end of the sequence. We also use a slot error rate
penalty, similarly to \newcite{wensclstm15}, to re-rank the sentences generated with
beam search. We use this method for all three of the LSTM, d-scLSTM, and hld-scLSTM models for fairness.

Similarly to the LSTM model, the d-scLSTM and hld-scLSTM generate delexicalized
sentences, \textit{i.e.}, they generate slot tokens instead of slot values directly. These slot tokens are replaced with slot values in a post-processing step which is a
fairly common step in task-oriented dialogue NLG literature.

\subsection{Evaluation}
In NLG tasks, improvements in automated metric scores are most commonly used to
demonstrate improvement in the generation task. However, these metrics have been
shown to only weakly correlate with human evaluation in the non task-oriented dialogue setting~\cite{LiuHowNotTo-D16-1230} and hence are not considered reliable measures
of improvement. Human evaluation is considered the metric of choice, but human
ratings are expensive to obtain. The ease of computing these automated metrics and their availability for rapid prototyping has lead to their widespread adoption.

We evaluate the models described in the previous section on the DSTC2 ~\cite{dstc2} and the Restaurants
datasets~\cite{restaurants} using these automated metrics. These datasets are some of the only available resources for studying NLG for task-oriented dialogue. The DSTC2 dataset contains dialogues between human users and a dialogue system in a restaurant domain. The dataset is annotated with dialogue acts, slot type, and slot values. The NLG component of the dialogue system used for data collection is templated. The Restaurants dataset was specifically proposed for NLG and provides, for a set of dialogue acts with slot types and slot values, two sentences generated by humans.

We present the results of our experiments in Table~\ref{table:word-overlap} and Table~\ref{table:embedding}.
The code for our automated metric evaluation pipeline is available at \url{https://github.com/Maluuba/nlg-eval}.
The scores of all the models on these automated metrics are very high. This indicates that there is significant word overlap between the generated and the reference sentences and that the NLG task on these datasets can be solved with a simple model such as the LSTM model. In effect, table~\ref{table:word-overlap} shows that the LSTM model performs comparably to the d-scLSTM model based on the word-overlap metrics. This can be explained by the fact that the d-scLSTM model has more parameters and might suffer from overfitting issues on these relatively small datasets.

The hld-scLSTM is considered to consistently outperform the other models based on the word-overlap metrics. As explained by \newcite{sharma2017nlg}, this improvement results from the model's access to the lexicalized slot values, due to which it can take into account the grammatical associations of the generated words near the output tokens, thereby generating higher quality sentences. However, Table~\ref{table:embedding} shows that sentence-embedding based metrics judge all the models except the random one to perform quite similarly with again, very high performance scores.  

In the next section, we add human evaluation for these models on these datasets.

\begin{figure*}
	\small
	\centering
	\includegraphics[width=\linewidth]{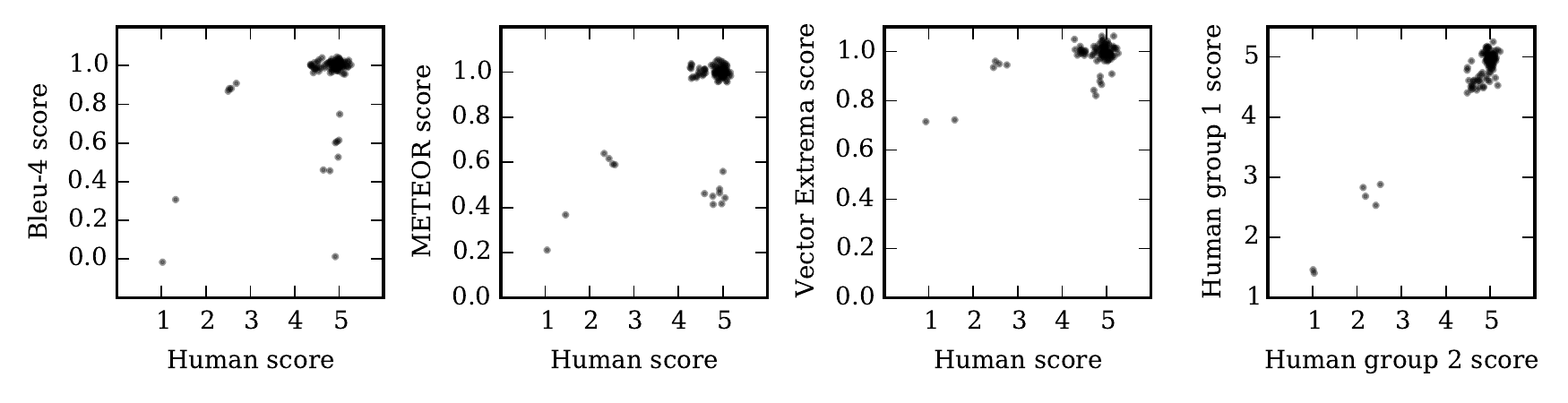}
	(a) DSTC2
	\includegraphics[width=\linewidth]{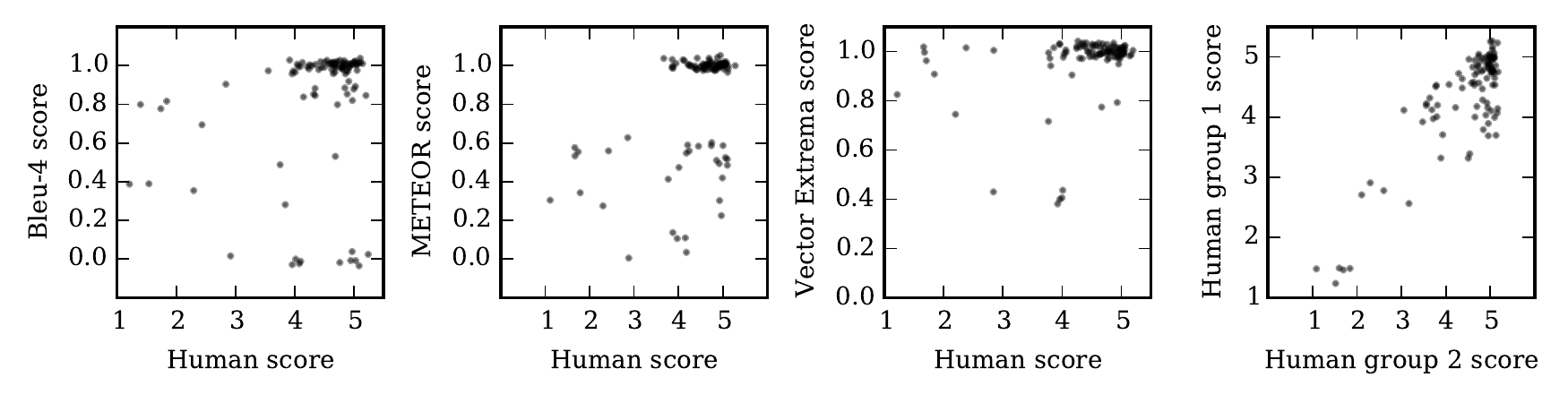}
	(b) Restaurants
	\caption{Scatter plots for correlation of some automated metrics with human evaluation for (a) the DSTC2 dataset, and (b) the Restaurants dataset. Random gaussian noise $\mathcal{N}(0, 0.1)$ has been added to data points along the human score axis and $\mathcal{N}(0, 0.02)$ has been added to the automated metric score's axis to aid visualization of overlapping data points. Transparency has been added for the same effect.}
	\label{fig:scatter-dstc2-rnnlg}
\end{figure*}

\subsection{Human rating collection}
\begin{table}
	\small
	\centering
	\begin{tabular}{|c|r|c|}
		\hline
		$\kappa$ & \# pairs & \% pairs \\ \hline
		\textgreater 0.1 & 55/55 & 100.0 \%\\
		\textgreater 0.2 & 40/55 &  72.7 \% \\
		\textgreater 0.3 & 28/55 &  50.9 \% \\
		\textgreater 0.4 & 19/55 &  34.5 \% \\
		\textgreater 0.5 &  8/55 &  14.5 \% \\
		\textgreater 0.6 &  0/55 &   0.0 \% \\ \hline
	\end{tabular}
	\caption{Pairwise Cohen's kappa scores for the 11 human users}
	\label{table:kappa}
\end{table}

We randomly selected 20 dialogue acts from the test set of each dataset. For
each of these contexts, we presented 5 sentences to the evaluators: the gold response provided in the
test set and the responses generated by the four models described in
Section~\ref{sec:models}. These sentences were randomly shuffled and not presented
in the same order. We invited 18 human users to score each of these 100 sentences
on a Likert-type scale of 1 to 5. The users were asked to rate the responses depending on how appropriate they were for the specified dialogue acts. A score of 1 was the lowest score, meaning that the response was not appropriate at all whereas a score of 5 meant that the sentence was highly appropriate.

We computed Cohen's kappa scores~\cite{cohenkappa} between the human users in pairs
of two. We removed 7 users who had kappa scores less than $0.1$ and used the
remaining 11 users for the correlation study. The kappa scores are presented in
Table~\ref{table:kappa}. Most of the user pairs have a Cohen's $\kappa > 0.3$
which indicates fair agreement between users~\cite{kappainterpret}.

\subsection{Correlation between automated metrics and human scores}
We present the correlation between the automated metrics and our collected human ratings
in Table~\ref{table:correlation}. We measure human \textit{v.s.} human correlation by randomly splitting
the human users into two groups. The results indicate that in most cases, human
scores correlate the best with other human scores. Except in the case of
the Spearman correlation for BLEU-N scores, we can see that there is a positive correlation between
the automated metrics and the human scores for these task-oriented datasets,  which
contrasts with the non task-oriented dialogue setting where
\newcite{LiuHowNotTo-D16-1230} observed no strong correlation trends. 

A likely explanation for the negative Spearman correlation values for BLEU-N is that there is only one gold reference per context in the DSTC2 dataset. The Restaurants dataset, on the other hand, provides two gold references per context.
Having multiple gold references increases the likelihood that the generated response will have significant word-overlap with one of the reference responses.

We present scatter plots for some of the metrics presented in Table~\ref{table:correlation}
in Figure~\ref{fig:scatter-dstc2-rnnlg}. We observe that all the
metrics correlate very well with humans on high scoring examples. As it can be seen in the scatter plots, most of the sentences are given the maximal score of 5 by the human evaluators. This confirms our previous observation that the available corpora for task-oriented dialogue NLG task are not very challenging and a simple LSTM-based model can output high-quality responses.

Overall, among the word overlap based automated metrics, METEOR consistently correlates with human evaluation on both datasets. These results confirm the original findings by \newcite{Banerjee:05} who showed that METEOR had good correlation with human evaluation in the machine translation task. The comparison with machine translation is highly relevant in the task-oriented setting because the NLG model essentially learns to translate the abstract representation of a sentence into a sentence. It is a translation task contrary to the non task-oriented setting where the NLG model needs to decide and output a new sentence based on the last sentence typed by a user and dialogue history. Therefore, automated metrics coming from the machine translation literature are more adequate in our case than in the non-task oriented case as shown by \newcite{LiuHowNotTo-D16-1230}.

It is interesting to see that METEOR correlates well with human evaluation consistently. This can be explained by the fact that even though METEOR does not rely on word embeddings, it includes notions of synonymy and paraphrasing when computing the alignment between the candidate and reference utterances. 

\section{Discussion}
We evaluated several natural language generation models trained on the DSTC2 and the Restaurants datasets based on several automated metrics. We also performed human evaluation on the model-generated responses and our study shows that human evaluation is a much more reliable metric compared to the others. Among the word-overlap based automated metrics, we found that the METEOR score correlates the most with human judgments and we suggest using METEOR for task-oriented dialogue natural language generation instead of BLEU. We also observe that these metrics are more reliable in the task-oriented dialogue setting compared to the general, non task-oriented one due to the limited possible diversity in the task-oriented setting. Also, as observed by \newcite{deltableu}, we can see that word-overlap based metrics correlate better with human evaluation when multiple references are provided, as in the Restaurants dataset. Otherwise, as in the case of DSTC2 which only provides one reference sentence per example, we observe that all the BLEU-N metrics negatively correlate with human evaluation on Spearman correlation.

As has been observed in the machine translation literature, using beam search improves the quality of generated sentences significantly compared to stochastic sampling. For similar models, our results show improvement in the automated metrics' scores compared to \newcite{wensclstm15} who used stochastic sampling for decoding instead of beam search.

\newcite{wensclstm15} did not use the slot error rate penalty with the vanilla LSTM model in their experiments. After adding the penalty in our case, we observe that the vanilla LSTM-based model performs as well as the delexicalized semantically-controlled LSTM model. This suggests that the added complexity introduced by the sc-LSTM unit does not offer a significant advantage for these two datasets.

High performance on automated metrics, achieved by our models on the DSTC2 and the Restaurants datasets lead us to conclude that these datasets are not very challenging for the NLG task. The task-oriented dialogue community should move towards using larger and more complex datasets, which have been recently announced, such as the Frames dataset~\cite{frames} or the E2E NLG Challenge dataset~\cite{e2enlg}.

\bibliography{acl2017}
\bibliographystyle{acl_natbib}

\end{document}